\documentclass[wcp]{jmlr}

\usepackage{longtable}

\usepackage{lineno}

\makeatletter
\let\Ginclude@graphics\@org@Ginclude@graphics 
\makeatother

\usepackage{bm,bbm,pifont}
\usepackage{multirow}
\usepackage{dcolumn}
\newcommand{\cmark}{\ding{51}}
\newcommand{\xmark}{\ding{55}}
\graphicspath{{figures/}}

\jmlrvolume{222}
\jmlryear{2023}
\jmlrworkshop{ACML 2023}

\title[A Pragmatic Look at Deep Imitation Learning]{A Pragmatic Look at Deep Imitation Learning}

\author{\Name{Kai Arulkumaran} \Email{kai\_arulkumaran@araya.org}\\
  \Name{Dan {Ogawa Lillrank}} \Email{dan\_ogawa@araya.org}\\
  \addr Araya Inc., Tokyo, Japan}

\begin{document}

\maketitle

\begin{abstract}
The introduction of the generative adversarial imitation learning (GAIL) algorithm has spurred the development of scalable imitation learning approaches using deep neural networks. Many of the algorithms that followed used a similar procedure, combining on-policy actor-critic algorithms with inverse reinforcement learning. More recently there have been an even larger breadth of approaches, most of which use off-policy algorithms. However, with the breadth of algorithms, everything from datasets to base reinforcement learning algorithms to evaluation settings can vary, making it difficult to fairly compare them. In this work we re-implement 6 different IL algorithms, updating 3 of them to be off-policy, base them on a common off-policy algorithm (SAC), and evaluate them on a widely-used expert trajectory dataset (D4RL) for the most common benchmark (MuJoCo). After giving all algorithms the same hyperparameter optimisation budget, we compare their results for a range of expert trajectories. In summary, GAIL, with all of its improvements, consistently performs well across a range of sample sizes, AdRIL is a simple contender that performs well with one important hyperparameter to tune, and behavioural cloning remains a strong baseline when data is more plentiful.
\end{abstract}

\section{Introduction}

Several years ago, \cite{henderson2018deep} formally brought attention to the reproducibility crisis in deep reinforcement learning (DRL). Some solutions have been to settle on evaluation protocols for common benchmarks \citep{machado2018revisiting}, improving the statistical tools with which we evaluate results \citep{agarwal2021deep}, or simply just providing reliable algorithm implementations \citep{raffin2021stable}. Still, sometimes it is necessary to take a step back and evaluate the (claimed) progress within different areas of machine learning \citep[ML;][]{oliver2018realistic,musgrave2020metric}. In on-policy RL it has already been observed that implementation details can matter more than the algorithmic contributions of novel algorithms \citep{engstrom2020implementation,andrychowicz2021matters}, and so in turn we have opted to take a \emph{pragmatic} look at (deep) imitation learning (IL), creating a single, open source codebase to fairly compare algorithms.\footnote{\url{https://github.com/Kaixhin/imitation-learning}}

IL is the branch of ML that is concerned with learning from ``demonstration'' data \citep{hussein2017imitation}. In other words, there exists agents, acting in environments, from which we collect data, in order to train our own agent. IL is intimately linked with RL---learning to act optimally in a given environment---and hence is often analysed with respect to concepts that are foundational to RL, such as states, actions, policies, reward functions, value functions, etc. \citep{sutton2018reinforcement}. Indeed, one of the most prominent approaches to IL is a technique known as inverse reinforcement learning \citep[IRL;][]{arora2021survey}.

In the same way that deep learning (DL) has enabled RL to scale to high-dimensional state and action spaces \citep{arulkumaran2017deep}, DL has also enabled IL to be applied to more complex domains. The most famous of these algorithms is generative adversarial IL \citep[GAIL;][]{ho2016generative}, which uses the generative adversarial framework \citep{goodfellow2014generative} to learn a reward function which can then be used with IRL. In their work, they show strong results against a range of baselines, including the simplest IL method, behavioural cloning \citep[BC;][]{pomerleau1988alvinn}. Although BC underperformed GAIL, they noted that it ``was able to reach satisfactory performance with enough data''---a common observation. However, to ``generate the datasets, ... [they] subsampled the expert trajectories''. While this reduces the number of samples available, it is not the same as providing fewer trajectories without subsampling, as the former tends to result in better coverage of the state space. \emph{Pragmatically}, collecting trajectory data is a bottleneck for IL, so subsampling should be avoided---and in this case BC's performance improves. IL results where trajectory data has been subsampled is not directly comparable to results where no subsampling has occurred, and unsurprisingly practitioners may miss critical details revealed in an appendix.\footnote{Never mind implementation details \emph{not} revealed within a paper.}

Another major issue in the reproducibility of IL algorithms is encountered in the GAIL paper---they generated their own expert trajectories using an RL agent. Out of the works that we review herein, nearly all generate their own expert data. And naturally, as time passes, the RL algorithms used within IL algorithms get replaced with more performant versions. One of the most significant of these changes is the move from on-policy RL algorithms, as used within the original work on GAIL, to more data-efficient off-policy algorithms \citep{kostrikov2019discriminator,blonde2019sample}. Do some of the on-policy IL algorithms that were competitive with GAIL \citep{kim2018imitation,wang2019random,brantley2020disagreement} still perform competitively when converted to be off-policy?

In this work, we review 6 different deep IL algorithms: GAIL \citep{ho2016generative}, generative moment matching imitation learning \citep[GMMIL;][]{kim2018imitation}, random expert distillation \citep[RED;][]{wang2019random}, disagreement-regularised imitation learning \citep[DRIL;][]{brantley2020disagreement}, adversarial reward-moment imitation learning \citep[AdRIL;][]{swamy2021moments}, and primal Wasserstein imitation learning \citep[PWIL;][]{dadashi2021primal}. To minimise differences between them, we update GMMIL, RED, and DRIL to be off-policy, give them access to absorbing state indicators \citep{kostrikov2019discriminator}, and use soft-actor critic \citep[SAC;][]{haarnoja2018soft} as the base RL algorithm for all methods. We then evaluate them on the standard MuJoCo continuous control benchmark environments \citep{todorov2012mujoco,brockman2016openai}, using the D4RL expert trajectory datasets \citep{fu2020d4rl} for reproducibility. For a fairer comparison, we give each algorithm the same hyperparameter optimisation budget, and then run them with the best settings for several seeds, and report results using current best practices \citep{agarwal2021deep}. Given the many improvements to GAIL, as well as prior effort performing extensive hyperparameter tuning on adversarial IL methods \citep{orsini2021matters}, unsurprisingly it remains one of the best IL algorithms to use. With more trajectories, AdRIL performs similarly to GAIL, whilst remaining simple to implement and tune (unlike GAIL). And, as observed many times before, BC becomes a competitive baseline with enough data.

\section{Background}

\subsection{Imitation Learning}

The goal of IL is to train a policy\footnote{In our case, a neural network with parameters $\theta$.}, $\hat{\pi}(a|s; \theta)$, mapping states $s$ to a distribution over actions $a$, to mimic an expert policy $\pi^*(a|s)$, given either the expert policy itself, or more commonly, a fixed dataset $\xi^* = \{\tau_1, \ldots, \tau_N\}$, of trajectories $\tau = \{s_0, a_0, s_1, a_1 , \ldots s_T, a_T\}$ generated by the expert, where $N$ denotes the number of expert trajectories provided.

A common assumption within IL is that both the expert and our agent inhabit a Markov decision process (MDP), defined by the tuple $(\mathcal{S}, \mathcal{A}, \mathcal{T}, \mathcal{R}, p_0, \gamma)$: $\mathcal{S}$ and $\mathcal{A}$ are the state and action spaces, ${\mathcal{T}: \mathcal{S} \times \mathcal{A} \rightarrow \mathcal{S}}$ is the state transition dynamics, ${\mathcal{R}: \mathcal{S} \times \mathcal{A} \rightarrow \mathbb{R}}$ is the reward function, $p_0(s)$ is the initial state distribution, and $\gamma \in [0, 1]$ is the discount factor (used to weight immediate vs. future rewards). The expert policy is optimal in the sense that ${\pi^* = \arg\!\max_{\pi \in \Pi}\mathbb{E}_{\tau \sim \pi}[R_0]}$, where the return at timestep $t$, $R_t$, is the discounted sum of rewards following a policy from state $s_t$ until the end of the episode at timestep $T$: ${R_t = \sum_{k=0}^{T-t}\gamma^k r_{t+k+1}}$. While in RL the goal is to interact with the environment in order to find $\pi^*$ \citep{sutton2018reinforcement}, in IL we do not have access to $\mathcal{R}$, and must instead find $\pi^*$ assuming that we have access to optimal trajectories.\footnote{In this work we do not consider the more complex settings that include suboptimal demonstrations and/or noisy observations.} All following methods, unless specified otherwise, can be implemented using neural networks, providing flexible function approximation that can scale to large state and/or action spaces.

\subsection{Reduction to Supervised Learning}

The simplest method, BC \citep{pomerleau1988alvinn}, reduces IL to a supervised learning problem. Using $a^*$ to denote the expert's actions, BC can be formulated as minimising the 1-step deviation from the expert trajectories:
\begin{align}
\arg\!\min_\theta \mathbb{E}_{s, a^* \sim \xi^*}[\mathcal{L}(a^*, \hat{\pi}(a|s; \theta))], \label{eq:bc}
\end{align}
\noindent where $\mathcal{L}$ can be, as in maximum likelihood estimation, the negative log likelihood.

BC is very simple, and benefits from a fixed objective over a stationary data distribution. However, as $\hat{\pi}$ is only trained on $s \sim \xi^*$, it can fail catastrophically when it diverges from the states covered by $\pi^*$. In order to mitigate this, $\hat{\pi}$ must be evaluated on the environment in order to correct for discrepancies between $s, a \sim \hat{\pi}$ and $s, a \sim \pi^*$.

Interactive IL methods solve this issue of compounding errors \citep{ross2011reduction} by iterating over running $\hat{\pi}$ in the environment, calculating $\pi^*(a|s)$ on $\hat{\pi}$'s state distribution, and using supervised learning on the new data \citep{daume2009search,ross2010efficient,ross2011reduction}. While these approaches solve the data distribution issue, they require access to an interactive expert during training, which may not be available in many scenarios.

\subsection{Inverse Reinforcement Learning}

IRL instead overcomes this distribution shift by using RL to train $\hat{\pi}$ to mimic $\pi^*$ in the environment. The procedure consists of iterating between the following two steps:
\begin{enumerate}
  \item Construct a reward function\footnote{In the parametric case, parameterised by $\phi$, but potentially nonparametric.} $\hat{\mathcal{R}}(s, a; \phi)$ using $\xi^*$, and optionally $\tau \sim \hat{\pi}$
  \item Train $\hat{\pi}$ using RL
\end{enumerate}


RL is more complicated than the typical supervised learning setting. In particular, as the policy evolves, the data distribution changes. 
In the case of IRL, $\hat{\mathcal{R}}$ changing over time can introduce further non-stationarity. 

The basic objective of IRL can be stated as:
\begin{align}
\arg\!\max_\theta\mathbb{E}_{\tau \sim \hat{\pi}(s, a; \theta)}[\hat{R}_0] \quad \text{such that} \quad \pi^* = \arg\!\max_{\pi \in \Pi}\mathbb{E}_{\tau \sim \pi}[\hat{R}_0],
\end{align}
\noindent where $\hat{R}$ is the return with respect to the learned reward function $\hat{\mathcal{R}}$. However, this is underspecified \citep{ng2000algorithms}; e.g., any policy is trivially optimal for $\hat{\mathcal{R}} = 0$. IRL algorithms therefore incorporate one or several of the following three properties.

Firstly, one can match the state-action distribution under $\pi^*$, known as the expert's occupancy measure ${\rho_{\pi^*} = \mathbb{E}_{\tau \sim \pi^*}\left[\sum_{t=0}^T\gamma^t \mathbbm{1}_{s, a}\right]}$ \citep{syed2008apprenticeship,ho2016generative}, or, alternatively, feature expectations \citep{ng2000algorithms}. This is achieved using the learned reward function, and is hence dependent on the expressivity of $\hat{\mathcal{R}}$. In particular, the constant function is underspecified and allows an infinite set of solutions. Secondly, one can ``penalise'' following trajectories taken by (previous iterations of) $\hat{\pi}$ \citep{ng2000algorithms}. This allows $\hat{\mathcal{R}}$ to focus on relevant parts of the state-action space, but implicitly assumes that current/past versions of $\hat{\pi}$ are suboptimal. Thirdly, one can use the maximum entropy principle \citep{jaynes1957information} to find a unique best solution out of the set of solutions that match the expert's occupancy measure/feature expectations \citep{ziebart2008maximum}. Using the Lagrangian multiplier $\lambda$, and denoting $H$ as the entropy, this results in the following modified RL objective: ${\arg\!\max_\theta\mathbb{E}_{\tau \sim \pi(s, a; \theta)}[R_0] + \lambda H[\pi(s, a; \theta)]}$. Entropy regularisation is a classic technique in RL \citep{williams1991function}.

A simple algorithm that uses these properties is soft Q imitation learning \citep[SQIL;][]{reddy2020sqil}. SQIL uses the constant reward function:
\begin{align}
\hat{\mathcal{R}} =
\begin{cases}
  1 & \text{if } (s, a) \in \xi^*\\
  0 & \text{if } (s, a) \sim \hat{\pi},
\end{cases} \label{eq:sqil}
\end{align}
\noindent which encourages not just imitating the expert's actions, but also visiting the same states. Building upon the maximum entropy model of expert behaviour \citep{ziebart2008maximum}, \cite{reddy2020sqil} show that their algorithm can be interpreted as regularised BC with a sparsity penalty on the reward function \citep{piot2014boosted}.\footnote{Due to $\hat{\mathcal{R}}$ being +1 at expert state-action pairs, and 0 elsewhere.} The full SQIL algorithm (for continuous action spaces) trains a SAC agent on half-half mixed batches of expert and agent data with its constant reward function. While simple, the downside of the constant reward function is that as the agent improves, its transitions still get labelled with zero rewards, potentially leading to a collapse in performance with over-training.

\subsection{Adversarial Imitation Learning}

Adversarial IL methods instead learn a reward function online using adversarial training, motivated by maximum entropy occupancy measure matching \citep{ho2016generative}. In generative adversarial network training \citep{goodfellow2014generative}, the ``generator'' is trained to output samples that fool the ``discriminator'' $D: \mathcal{S} \times \mathcal{A} \rightarrow (0, 1)$, whilst the discriminator is trained to discriminate between samples from the generator and the data distribution. This is a minimax game, in which the equilibrium solution corresponds to minimising the Jensen-Shannon divergence between the generated and real distributions. In GAIL, $\hat{\pi}$ plays the role of the generator, and the discriminator is trained on state-action pairs from $\hat{\pi}$ and $\pi^*$: ${\min_G \max_D \mathbb{E}_{s, a \sim \pi^*}[\log(D(s, a))] + \mathbb{E}_{s, a \sim \hat{\pi}}[\log(1 - D(s, a))]}$. Under this formulation, higher values indicate how ``expert'' $D$ believes its input to be.

There are several options for constructing $\hat{\mathcal{R}}$ from $D$. Prominent examples include those introduced in GAIL, adversarial inverse reinforcement learning \citep[AIRL;][]{fu2018learning} (corresponding to the reverse Kullback-Leibler (KL) divergence $D_{\text{KL}}(\rho_{\hat{\pi}} \Vert \rho_{\pi^*})$), and forward KL AIRL \citep[FAIRL;][]{ghasemipour2020divergence} (Table \ref{tab:adv_reward_fns}). As discussed by \cite{kostrikov2019discriminator} and empirically investigated by \cite{jena2020addressing}, there is a potential reward bias in these functions. They note that positive $\hat{\mathcal{R}}$, i.e., $-\log(1 - D(s, a))$, biases agents towards survival, whereas negative $\hat{\mathcal{R}}$, i.e., $\log(D(s, a))$ biases agents towards early termination. This bias means that even constant reward functions can outperform either of these depending on the type of the environment. We recommend the original works for discussions on the properties of various reward functions \citep{kostrikov2019discriminator,jena2020addressing,ghasemipour2020divergence}. \cite{kostrikov2019discriminator} also make the observation that many IL algorithms do not correctly handle terminal states, and propose appending an absorbing state indicator to states, which allows IRL algorithms to properly estimate values for terminal states. This requires processing complete trajectories from both the expert and the agent, and allowing both the RL agent and the discriminator to learn from the indicator feature.

\begin{table}
  \caption{Adversarial imitation reward functions \citep{ghasemipour2020divergence}.} 
  \label{tab:adv_reward_fns}
  \centering
  \resizebox{\columnwidth}{!}{
  \begin{tabular}{lccc}
    \hline
    & $\hat{\mathcal{R}}$ & Positive (bounded) & Negative (bounded)\\
    \hline
    GAIL & $\log D(s, a)$ & \xmark (-) & \cmark (\xmark)\\
    AIRL & $h(s, a) = \log(D(s, a)) - \log(1 - D(s, a))$ & \cmark (\xmark) & \cmark (\xmark)\\
    FAIRL & $-h(s, a) \cdot e^{h(s, a)}$ & \cmark (\cmark) & \cmark (\xmark)\\
    \hline
  \end{tabular}
  }
\end{table}

While GAIL implicitly returns a reward function, if trained to optimality then $D$ will return 0.5 for state-action pairs from both $\hat{\pi}$ and $\pi^*$. \cite{finn2016connection} propose changing the form of the $D$ to $\frac{\exp(f(\tau))}{\exp(f(\tau)) + \hat{\pi}(\tau)}$, allowing the optimal reward function to be recovered as $f (+ \text{const})$. AIRL makes a practical algorithm from this by changing $D$ to operate over state-action pairs, as in GAIL, and also further disentangling the recovered reward function $f$ as the sum of a reward approximator $g(s, a)$ and a reward shaping term \citep{ng1999policy} $h(s)$: ${f(s, a, s') = g(s, a) + \gamma h(s') - h(s)}$, where $s'$ is the successor state.

One of the most significant improvements to adversarial IL methods came from moving to more sample-efficient off-policy RL algorithms \citep{kostrikov2019discriminator,blonde2019sample}, which perform updates on batches of data stored in an experience replay memory \citep{lin1992self}. The discriminator can similarly be more efficiently trained on replay data, and although this should include an importance weighting term to account for the change in data distribution, in practice this is not needed \citep{kostrikov2019discriminator}.

There are countless more advances within adversarial IL, making it difficult to know which techniques increase performance robustly. \cite{orsini2021matters} performed a large-scale hyperparameter search over many of these methods. The key takeaways were that off-policy RL algorithms help improve sample efficiency, discriminator regularisation is key, and that hyperparameter choices which are optimal for AI-generated trajectories are not always the same for human-generated trajectories---a valuable distinction that lies out of the scope of this work. Their work also shows the importance of large-scale empirical evaluation, as their results overturned theoretical claims about the importance of discriminator regularisation \citep{blonde2022lipschitzness}.

\subsection{Distribution Matching Imitation Learning}

One disadvantage of adversarial training is the requirement for the discriminator, which is also undergoing training as part of the minimax game, to provide a useful training signal to the generator. There are several other IRL algorithms that also attempt to match the expert and agent's state-action distributions, but use non-adversarial methods.

One solution is to replace the discriminator with a nonparametric model \citep{li2015generative,dziugaite2015training}. Specifically, distribution matching can be achieved by minimising the maximum mean discrepancy \citep[MMD;][]{gretton2012kernel} defined over a reproducing kernel Hilbert space (RKHS). Given distributions, $P$ and $Q$, and a mapping ${\psi: \mathcal{X} \rightarrow \mathcal{H}}$ from features $X \in \mathcal{X}$ to an RKHS $\mathcal{H}$, the MMD is the distance between the mean embeddings of the features: ${\text{MMD}(P, Q) = \Vert\mathbb{E}_{x \sim P}[\psi(x)] - \mathbb{E}_{y \sim Q}[\psi(y)]\Vert_\mathcal{H}}$. Using a kernel function $k$, one can calculate $\text{MMD}^2(P, Q) = \mathbb{E}_{x, x' \sim P} k(x, x') + \mathbb{E}_{y, y' \sim Q} k(y, y') - 2\mathbb{E}_{x \sim P, y \sim Q} k(x, y)$.

GMMIL \citep{kim2018imitation} extends this principle to the IL setting. Dropping terms that are constant with respect to $\hat{\pi}$, GMMIL has the reward function:
\begin{align}
\hat{\mathcal{R}} = \frac{1}{M}\sum_{i=1}^M k((s, a), (s_i^*, a_i^*)) - \frac{1}{N}\sum_{j=1}^N k((s, a), (s_j, a_j)), \label{eqn:gmmil}
\end{align}
where $M$ and $N$ are the number of state-action pairs from $\pi^*$ and $\hat{\pi}$, respectively.

Two disadvantages of GMMIL are that 1) the ``discriminator'' cannot learn relevant features, and 2) it has $O(MN)$ complexity. RED \citep{wang2019random} solves these issues by building upon random network distillation \citep[RND;][]{burda2018exploration}. In RND, a predictor network ${f_\phi: \mathcal{S} \times \mathcal{A} \rightarrow \mathbb{R}^K}$ is trained to minimise the mean squared error (MSE) against a fixed, randomly initialised network ${f_{\bar{\phi}}: \mathcal{S} \times \mathcal{A} \rightarrow \mathbb{R}^K}$. Empirically, the MSE indicates how out-of-distribution new data is. RED uses a Gaussian function over the MSE, resulting in
\begin{align}
\hat{\mathcal{R}} = \exp(-\sigma \Vert f_\phi(s, a) - f_{\bar{\phi}}(s, a)\Vert_2^2),
\end{align}
where $\sigma$ is a bandwidth hyperparameter. \cite{wang2019random} interpret RND as an approximate support estimation method, and hence the RED reward function encourages the agent to have a support over its state-action distribution that matches the expert's.

Similarly to RED, DRIL \citep{brantley2020disagreement} constructs a reward function based on the disagreement between models trained on the expert data, and can also be interpreted as a support estimation method. However, unlike the other methods which operate over the joint distribution of state-action pairs, DRIL builds simply upon BC, operating over $p(a|s)$. DRIL first trains an ensemble of $E$ different policies using the BC objective (Equation \ref{eq:bc}) on the expert data, and then uses a function of the (negative of the) variance between the policies to estimate a reward for the agent:
\begin{align}
\hat{\mathcal{R}} &= -C_\text{U}^\text{clip}(s, a) =
\begin{cases}
  1 & \text{if } \text{Var}_{\pi \in \Pi_E}[\pi(a|s)] \leq q\\
  -1 & \text{otherwise},
\end{cases} \label{eq:dril}
\end{align}
where the $q$ is a top quantile of the uncertainty cost computed over the expert dataset. While deep ensembles are known to produce reasonable uncertainty estimates (i.e., variance in outputs) on out-of-distribution data \citep{lakshminarayanan2017simple}, it is also possible to approximate them using sampling with dropout \citep{srivastava2014dropout}. \cite{brantley2020disagreement} showed empirically that this performed comparatively to using independent models.

The Wasserstein distance is another way of defining a distance between two probability distributions on a given metric space $\mathcal{M}$, and minimising it can be interpreted as finding the optimal coupling, $\gamma$, for transporting probability mass from one distribution to other, whilst minimising the transport cost given by a metric $d$ on $\mathcal{M}$ \citep{villani2009optimal}.\footnote{Note that in this subsection alone we use $\gamma$ for couplings, in line with optimal transport literature.} PWIL \citep{dadashi2021primal} aims to minimise the Wasserstein-2 distance between the agent and expert's state-action distributions:
\begin{align}
\inf_{\pi \in \Pi} \mathcal{W}_2^2(\rho_{\hat{\pi}}, \rho_{\pi^*}) = \inf_{\pi \in \Pi} \inf_{\gamma \in \Gamma} \sum_{t=1}^T \underbrace{\sum_{m=1}^M d((s_t, a_t), (s_t^*, a_t^*))^2\gamma[t, m]}_{c_{t,\pi}},
\end{align}
\noindent where $c_{t,\pi}$ is the (time-dependent) optimal transport cost.

The optimal coupling for policy $\pi$, $\gamma_\pi^*$, requires the full trajectory generated by $\pi$, so \cite{dadashi2021primal} define a greedy coupling $\gamma_\pi^g$ that transports probability mass at each timestep $t$, allowing the cost to be calculated online as the agent interacts with the environment. The cost with the greedy coupling, $c_{t,\pi}^g$, is an upper bound to the Wasserstein distance, and hence optimising it still minimises the distance between the agent and expert's state-action distribution. The reward can be defined by applying a monotonously decreasing function to the cost:
\begin{align}
\hat{\mathcal{R}_t} = \alpha \exp(-\frac{\beta T}{\sqrt{|\mathcal{S}| + |\mathcal{A}|}}c_{t,\pi}^g),
\end{align}
\noindent where $\alpha$ and $\beta$ are reward scale hyperparameters, and $d$ is set to the Euclidean distance between Z-score normalised state-action pairs.

Another view on distribution matching, known as moment matching\footnote{Matching the moments of the model distribution to the empirical target distribution.}, can be achieved through optimising integral probability metrics \citep[IPM;][]{muller1997integral}. IPMs provide a distance function between two distributions, $\sup_{f \in \mathcal{F}} \mathbb{E}_{x \sim P}[f(x)] - \mathbb{E}_{y \sim Q}[f(y)]$, for a function class $\mathcal{F}$ containing functions $f: \mathcal{X} \rightarrow \mathbb{R}$. Different function classes $\mathcal{F}$ recover different IPMs; for instance, $\mathcal{F} = \{f: \Vert f \Vert_\mathcal{H} \leq 1\}$ with RKHS $\mathcal{H}$ gives the MMD and $\mathcal{F} = \{f: \Vert f \Vert_L \leq K\}$ with bounded Lipschitz
constant $K$ gives the Wasserstein distance.

\cite{swamy2021moments} use this to provide a more general view on IL, arguing that training an agent to match the moments of the expert's reward or action-value distributions will achieve the same performance. When the agent is able to interact with the environment, \cite{swamy2021moments} show that it is possible to do reward moment-matching, with a form that is similar to GMMIL, penalising the difference in moments between the agent and expert's state-action pairs. However, in their view on IL they also focus on the moment matching happening within a minimax game between the agent and the reward function, which motivates the need to update the reward function. Solving for a closed-form reward function in an RKHS with the indicator kernel function, the AdRIL reward function is:
\begin{align}
\hat{\mathcal{R}} =
\begin{cases}
  \frac{1}{|\xi^*|} & \text{if } (s, a) \in \xi^*\\
  0 & \text{if } (s, a) \sim \hat{\pi}\\
  \frac{-1}{\text{round}\cdot|\xi|} & \text{if } (s, a) \sim \hat{\pi}_\text{old},
\end{cases} \label{eq:adril}
\end{align}
\noindent where the final term assigns a negative reward to state-action pairs from old trajectories, inversely proportional to the number of rounds of updates (a hyperparameter that corresponds to a fixed number of updates), and the current number of agent trajectories, $|\xi|$. AdRIL can therefore be considered an improvement upon SQIL's constant reward function, obviating the need for early stopping \citep{reddy2020sqil}.

\section{Experiments}
\label{sec:experiments}

\subsection{Environments + Data}

We evaluate all algorithms on the popular MuJoCo simulated robotics benchmarks: Ant, HalfCheetah, Hopper, and Walker2D \citep{todorov2012mujoco,brockman2016openai}. To improve reproducibility and enable fairer comparison against other reported results, we use the D4RL ``expert-v2'' trajectory datasets \citep{fu2020d4rl}.\footnote{Although this benchmark was developed for offline RL, we use it for IL by ignoring the saved rewards.} When loading the expert data we process each episode to distinguish between ``true'' and time-dependent terminations \citep{pardo2018time}, and provide absorbing state indicators \citep{kostrikov2019discriminator}; these are also tracked for agent episodes. By default, we maximise available data by not subsampling expert transitions. We choose 3 different trajectory ``budgets'' for the IL algorithms to learn from: 5, 10 and 25 expert trajectories.

\subsection{Algorithms + Hyperparameter Search + Evaluation}

All IL algorithms use SAC \citep{haarnoja2018soft}, with automatic entropy tuning \citep{haarnoja2018softa}, as the base RL agent, as theoretically required by SQIL and AdRIL, and as was empirically shown to be performant for adversarial IL algorithms \citep{orsini2021matters}. The actor applies a tanh transformation to scale actions $\in (-1, 1)$, and dual critics are trained to reduce value overestimation \citep{fujimoto2018addressing}. We also include BC as a baseline, with the same actor architecture. All algorithms are optimised with AdamW \citep{loshchilov2019decoupled}. We use PyTorch \citep{paszke2019pytorch} for all of our code.

We group the IL algorithms and their variants into 6 key methods: AdRIL, DRIL, GAIL, GMMIL, PWIL, and RED. Our hyperparameter search spaces were determined based on hyperparameter ranges within the original works, a large subset of options tried by \cite{orsini2021matters}, and general hyperparameters such as learning rate and batch size, resulting in 7-18 hyperparameters to tune per algorithm. For each trajectory budget, we give each algorithm 30 hyperparameter evaluations using Bayesian optimisation \citep{balandat2020botorch}, with the minimum of the cumulative reward for each of the 4 environments used as the optimisation objective, which can be seen as minimising regret over a set of environments. We then evaluate agents over 10 seeds with the best hyperparameters found, and report performance according to best practices \citep{agarwal2021deep}. We therefore train 2880 agents for the final results, not including the extensive training and testing of agents performed while replicating and augmenting IL algorithms.

DRIL, GAIL and RED include several options for their trained discriminators, including network hidden size, depth, activation function, dropout, and weight decay. The GAIL discriminator has additional options, detailed below.

AdRIL options include balanced sampling \citep[alternating sampling expert and agent data batches vs. mixed batches,][]{swamy2021moments}, and the discriminator update frequency $\geq 1$, which determines the number of ``rounds''. We also include ``0'' in the search space, which if chosen reverts to using the SQIL reward function.

DRIL options include the quantile cutoff $\in [0, 1]$. As per the original work \citep{brantley2020disagreement}, we include an auxiliary BC loss during training. For simplicity we use the dropout ensemble. As using absorbing state indicators requires importance sampling, we adapt the BC loss used within DRIL to account for importance weights.

GAIL options include reward shaping \citep{fu2018learning}, subtracting $\log\hat{\pi}(a|s)$ from the predicted reward \citep{fu2018learning}, the GAIL, AIRL and FAIRL reward functions \citep{ho2016generative,fu2018learning,ghasemipour2020divergence}, discriminator gradient penalty \citep{kostrikov2019discriminator,blonde2019sample}, discriminator spectral normalisation \citep{blonde2022lipschitzness}, discriminator entropy bonus $\geq 0$ \citep{orsini2021matters}, binary cross-entropy, Mixup, and nn-PUGAIL discriminator loss functions \citep{ho2016generative,chen2021batch,xu2021positive}, and 3 additional hyperparameters for these loss functions (Mixup alpha $\geq 0$, positive class prior $\geq 0, \leq 1$, and non-negative margin $\geq 0$).

To adapt GMMIL for absorbing states, we adapt it to use the weighted MMD \citep{yan2017mind}. Due to the $O(MN\cdot\dim(X))$ complexity of MMD, it is prohibitive to use the entire expert dataset per update in the off-policy setting, and hence we randomly sample a batch of expert transitions per update. For this reason we also restrict the maximum batch size of GMMIL's hyperparameter search space.

PWIL options include the reward scale $\alpha \geq 0$ and reward bandwidth scale $\beta \geq 0$. To make PWIL more comparable to the other algorithms, we use the ``nofill'' variant, which does not prefill the replay buffer with expert transitions \citep{dadashi2021primal}.

To adapt RED for absorbing states, we train its discriminator with a weighted MSE loss.

The hyperparameter search spaces and optimised hyperparameters can be found documented in the codebase.

\subsection{Results}

\begin{figure}
  \centering
  \includegraphics[width=\textwidth]{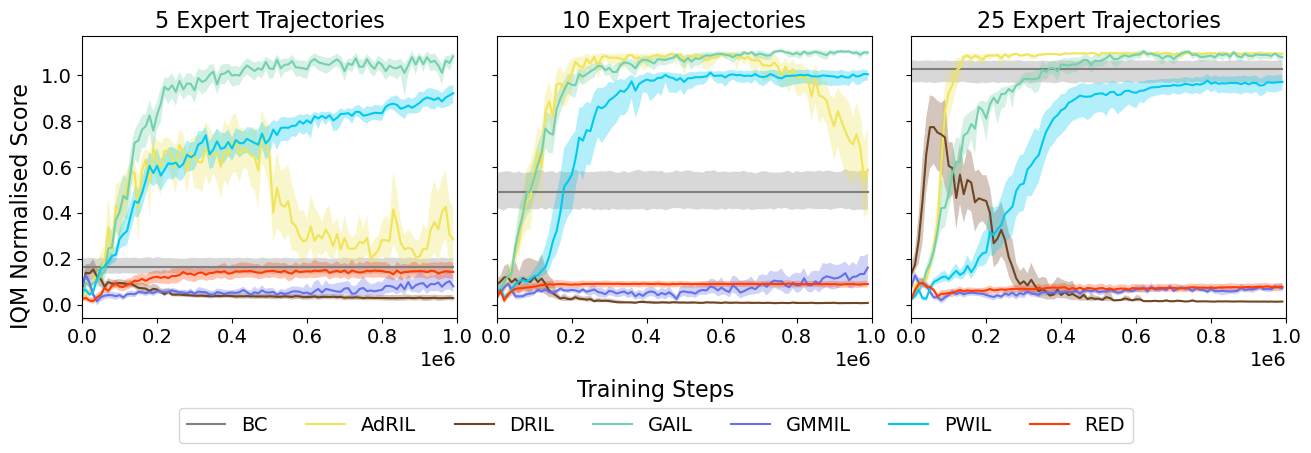}
  \caption{Normalised scores over different trajectory budgets (IQM ± 95\% CI). GAIL performs well across all budgets, with AdRIL performing joint-best for high budgets. AdRIL's weakness is its sensitivity to the discriminator update frequency, which can cause policy collapse if not tuned well. PWIL is reasonably strong across budgets. BC is a strong baseline for high budgets.}
  \label{fig:results}
\end{figure}

All IL algorithms (except BC) are trained with $10^6$ environment interaction steps, and are evaluated 30 times with a deterministic policy every $10^4$ steps. To aggregate test returns, we calculate the interquartile mean (IQM) over the 30 evaluation episodes, and then once again over seeds, reporting the final IQM ± 95\% confidence interval (CI) using 1000 stratified bootstrap samples \citep{agarwal2021deep}. The final returns over all environments are reported in Table \ref{tab:results}.\footnote{For reference we also include the results of our SAC implementation, trained for $3 \times 10^6$ steps.} We also show performance over time in Figure \ref{fig:results}, where the returns are aggregated over environments, normalised by the D4RL environment min and max reference scores: $\text{normalised score} = \frac{\text{score - random score}}{\text{expert score - random score}}$, with the reference scores produced by random and expert agents.

\begin{table}
  \caption{Returns over different trajectory budgets and environments (IQM ± 95\% CI). Optimising for minimum regret across all environments results in hyperparameter choices that are suboptimal for individual environments. Although GAIL performs the best over all trajectory budgets for most environments, AdRIL is always the most effective in the Hopper environment.} 
  \label{tab:results}
  \centering
  \resizebox{\columnwidth}{!}{
  \begin{tabular}{clD{,}{\ \pm\ }{7,7}D{,}{\ \pm\ }{7,7}D{,}{\ \pm\ }{7,7}D{,}{\ \pm\ }{7,7}}
    \hline
    \# Trajectories & & \multicolumn{1}{c}{Ant} & \multicolumn{1}{c}{HalfCheetah} & \multicolumn{1}{c}{Hopper} & \multicolumn{1}{c}{Walker2D}\\
    \hline
    & SAC & 5155.97,812.74 & 13994.64,2304.52 & 2086.81,608.70 & 5903.79,340.09\\
    \hline
    \multirow{8}{1em}{5} & BC & 587.82,255.80 & 52.48,153.96 & 701.03,134.96 & 909.28,558.42\\
    & AdRIL & 1191.88,787.27 & 36.58,60.07 & \textbf{3584.20},\textbf{30.61} & 169.10,576.42\\
    & DRIL & -89.15,84.87 & -63.48,20.80 & 367.35,324.27 & -20.69,58.24\\
    & GAIL & \textbf{5439.19},\textbf{253.85} & \textbf{11238.76},\textbf{217.55} & 3517.76,178.49 & \textbf{5110.03},\textbf{66.72}\\
    & GMMIL & 575.01,903.94 & 299.80,272.75 & 820.35,488.22 & 60.05,56.30\\
    & PWIL & 3168.17,1285.61 & 10775.81,386.64 & 3529.61,32.47 & 3678.89,167.96\\
    & RED & -90.94,252.32 & -111.72,466.09 & 2500.67,1024.09 & 1052.36,184.32\\
    \hline
    \multirow{8}{1em}{10} & BC & 2178.80,311.20 & 184.06,176.27 & 1139.22,559.33 & 4687.82,380.03\\
    & AdRIL & 115.84,520.84 & 5957.19,2747.32 & \textbf{3594.14},\textbf{8.50} & 2267.06,2121.80\\
    & DRIL & -318.65,104.92 & -117.86,32.44 & 37.51,103.46 & -12.32,8.58\\
    & GAIL & \textbf{5368.46},\textbf{54.49} & \textbf{11571.12},\textbf{154.39} & 3580.90,85.99 & \textbf{5041.87},\textbf{51.92}\\
    & GMMIL & 590.47,1022.66 & 448.40,643.33 & 1325.33,307.86 & 235.79,196.27\\
    & PWIL & 5133.04,72.93 & 10591.22,1204.07 & 3558.52,224.55 & 4185.11,490.59\\
    & RED & 6.15,168.53 & 978.77,408.87 & 322.68,62.43 & 216.50,197.14\\
    \hline
    \multirow{8}{1em}{25} & BC & 4698.29,207.67 & 5550.51,1780.47 & 3271.88,369.44 & 4955.85,21.21\\
    & AdRIL & 5246.33,70.14 & 11244.16,129.42 & \textbf{3579.96},\textbf{12.15} & 4962.86,15.00\\
    & DRIL & 1077.66,475.96 & -60.11,27.60 & 9.24,1.16 & -29.86,10.68\\
    & GAIL & \textbf{5557.74},\textbf{81.01} & \textbf{11392.09},\textbf{377.15} & 3467.73,299.07 & \textbf{5055.76},\textbf{64.13}\\
    & GMMIL & -7.81,176.54 & 620.01,319.07 & 854.59,244.31 & 13.29,48.66\\
    & PWIL & 5040.18,114.00 & 10785.65,170.43 & 3418.18,434.71 & 3855.43,296.41\\
    & RED & 284.14,181.12 & -62.76,181.27 & 323.86,22.49 & 246.23,165.69\\
    \hline
  \end{tabular}
  }
\end{table}

Overall, GAIL performs robustly across all trajectory budgets, presumably due to the large amount of research that has gone into improving the performance and robustness of adversarial IL methods. However, AdRIL, which is practically much simpler (and faster), performs the same (with higher sample efficiency) with higher budgets, as near-expert data from the agent is less likely to overpower the effect of the expert data. This can also be seen in the hyperparameter optimisation, with the discriminator update frequency going from 0 (reverting to the SQIL reward function) to 12500 to 25000 as the trajectory budget increased. PWIL also performed well across budgets. Hyperparameter optimisation set the reward ($\alpha$) and reward bandwidth ($\beta$) scales to 1 for 5 trajectories, but increased the values for these for 10 and 25 trajectories. However, we note that since each environment step requires computing the reward between the current state-action pair and a subset of the expert data, starting with the entire dataset at the start of each episode, it is computationally expensive for high budgets. BC scales well as the number of trajectories increase.

Unfortunately, we were unable to successfully optimise off-policy versions of DRIL, GMMIL and RED. In an early version of our codebase we were able to optimise their original, on-policy versions successfully, so with considerable effort put into hyperparameter tuning and trying additional regularisation strategies, we believe that their is some fundamental issue caused from going from training on on-policy to off-policy returns. One would expect that for successful training the inferred rewards for the agent's trajectories should increase over time, but this was observed for runs of these methods as well, and is therefore not predictive of successful imitation. Q-values are a function of the predicted rewards, so did not provide further diagnostic insights. We also created variants of DRIL and RED in which the discriminators were trained online, similarly to GAIL, but were unsuccessful; however, there are many ways to do so, and our attempts do not preclude a successful online discriminator variant from being developed. Finally, we note that DRIL performs better at the start of training, which we can attribute to the BC auxiliary loss; experiments with uncertainty-only DRIL \citep[UO-DRIL;][]{brantley2020disagreement} did not show this trend, with scores remaining low during the entirety of training.

Some weak trends we noticed from hyperparameter optimisation were that both batch and discriminator sizes increased with the trajectory budget. We hypothesise that the former is due to added stochasticity in optimisation aiding when data is scarce, whilst the latter is due to the need to prevent overfitting in low-data regimes. However, we caution that these trends do not always hold, as, for example, the optimal batch size for GAIL decreased with the trajectory budget.

\section{Discussion}

In this paper, we took a pragmatic look at deep IL methods, reviewing the relationships between the different approaches, updated older methods to use more data-efficient off-policy RL algorithms, and finally performed a fair comparison between them on a standard benchmark. As BC is simple and does not involve environment interaction, we recommend that it should always be considered as a baseline. AdRIL is an attractive option for deep IL due to its simplicity and strong performance, although it has one critical hyperparameter that needs tuning. And although the myriad of options for GAIL make it more complicated to work with, we have empirical data on what does and doesn't work \citep{orsini2021matters}.

Although we were only able to test extensively on standard environments with expert data, we plan to release our framework to enable further, fair experiments on different environments, datasets, algorithms. Valuable open questions in the field of IL remain in the use of proxy reward functions for evaluating IL \citep{hussenot2021hyperparameter}, and how best to learn from human demonstration data \citep{orsini2021matters}.

\acks{This work was supported by JST, Moonshot R\&D Grant Number JPMJMS2012.}

\bibliography{main}

\begin{thebibliography}{62}
\providecommand{\natexlab}[1]{#1}
\providecommand{\url}[1]{\texttt{#1}}
\expandafter\ifx\csname urlstyle\endcsname\relax
  \providecommand{\doi}[1]{doi: #1}\else
  \providecommand{\doi}{doi: \begingroup \urlstyle{rm}\Url}\fi

\bibitem[Agarwal et~al.(2021)Agarwal, Schwarzer, Castro, Courville, and
  Bellemare]{agarwal2021deep}
Rishabh Agarwal, Max Schwarzer, Pablo~Samuel Castro, Aaron~C Courville, and
  Marc Bellemare.
\newblock {Deep Reinforcement Learning at the Edge of the Statistical
  Precipice}.
\newblock In \emph{NeurIPS}, 2021.

\bibitem[Andrychowicz et~al.(2021)Andrychowicz, Raichuk, Sta{\'n}czyk, Orsini,
  Girgin, Marinier, Hussenot, Geist, Pietquin, Michalski,
  et~al.]{andrychowicz2021matters}
Marcin Andrychowicz, Anton Raichuk, Piotr Sta{\'n}czyk, Manu Orsini, Sertan
  Girgin, Raphael Marinier, L{\'e}onard Hussenot, Matthieu Geist, Olivier
  Pietquin, Marcin Michalski, et~al.
\newblock {What Matters in On-policy Reinforcement Learning? A Large-scale
  Empirical Study}.
\newblock In \emph{ICLR}, 2021.

\bibitem[Arora and Doshi(2021)]{arora2021survey}
Saurabh Arora and Prashant Doshi.
\newblock {A Survey of Inverse Reinforcement Learning: Challenges, Methods and
  Progress}.
\newblock \emph{Artif. Intell.}, 297:\penalty0 103500, 2021.

\bibitem[Arulkumaran et~al.(2017)Arulkumaran, Deisenroth, Brundage, and
  Bharath]{arulkumaran2017deep}
Kai Arulkumaran, Marc~Peter Deisenroth, Miles Brundage, and Anil~Anthony
  Bharath.
\newblock {Deep Reinforcement Learning: A Brief Survey}.
\newblock \emph{IEEE SPM}, 34\penalty0 (6):\penalty0 26--38, 2017.

\bibitem[Balandat et~al.(2020)Balandat, Karrer, Jiang, Daulton, Letham, Wilson,
  and Bakshy]{balandat2020botorch}
Maximilian Balandat, Brian Karrer, Daniel Jiang, Samuel Daulton, Ben Letham,
  Andrew~G Wilson, and Eytan Bakshy.
\newblock {BoTorch: A Framework for Efficient Monte-Carlo Bayesian
  Optimization}.
\newblock In \emph{NeurIPS}, 2020.

\bibitem[Blond{\'e} and Kalousis(2019)]{blonde2019sample}
Lionel Blond{\'e} and Alexandros Kalousis.
\newblock {Sample-efficient Imitation Learning via Generative Adversarial
  Nets}.
\newblock In \emph{AISTATS}, 2019.

\bibitem[Blond{\'e} et~al.(2022)Blond{\'e}, Strasser, and
  Kalousis]{blonde2022lipschitzness}
Lionel Blond{\'e}, Pablo Strasser, and Alexandros Kalousis.
\newblock {Lipschitzness is All You Need to Tame Off-policy Generative
  Adversarial Imitation Learning}.
\newblock \emph{Mach. Learn.}, 111\penalty0 (4):\penalty0 1431--1521, 2022.

\bibitem[Brantley et~al.(2020)Brantley, Sun, and
  Henaff]{brantley2020disagreement}
Kiant{\'e} Brantley, Wen Sun, and Mikael Henaff.
\newblock {Disagreement-regularized Imitation Learning}.
\newblock In \emph{ICLR}, 2020.

\bibitem[Brockman et~al.(2016)Brockman, Cheung, Pettersson, Schneider,
  Schulman, Tang, and Zaremba]{brockman2016openai}
Greg Brockman, Vicki Cheung, Ludwig Pettersson, Jonas Schneider, John Schulman,
  Jie Tang, and Wojciech Zaremba.
\newblock {OpenAI Gym}.
\newblock \emph{arXiv:1606.01540}, 2016.

\bibitem[Burda et~al.(2018)Burda, Edwards, Storkey, and
  Klimov]{burda2018exploration}
Yuri Burda, Harrison Edwards, Amos Storkey, and Oleg Klimov.
\newblock {Exploration by Random Network Distillation}.
\newblock In \emph{ICLR}, 2018.

\bibitem[Chen et~al.(2021)Chen, Nam, Nair, and Finn]{chen2021batch}
Annie~S Chen, HyunJi Nam, Suraj Nair, and Chelsea Finn.
\newblock {Batch Exploration with Examples for Scalable Robotic Reinforcement
  Learning}.
\newblock \emph{IEEE RA-L}, 6\penalty0 (3):\penalty0 4401--4408, 2021.

\bibitem[Dadashi et~al.(2021)Dadashi, Hussenot, Geist, and
  Pietquin]{dadashi2021primal}
Robert Dadashi, Leonard Hussenot, Matthieu Geist, and Olivier Pietquin.
\newblock {Primal Wasserstein Imitation Learning}.
\newblock In \emph{ICLR}, 2021.

\bibitem[Daum{\'e} et~al.(2009)Daum{\'e}, Langford, and Marcu]{daume2009search}
Hal Daum{\'e}, John Langford, and Daniel Marcu.
\newblock {Search-based Structured Prediction}.
\newblock \emph{Mach. Learn.}, 75\penalty0 (3):\penalty0 297--325, 2009.

\bibitem[Dziugaite et~al.(2015)Dziugaite, Roy, and
  Ghahramani]{dziugaite2015training}
Gintare~Karolina Dziugaite, Daniel~M Roy, and Zoubin Ghahramani.
\newblock {Training Generative Neural Networks via Maximum Mean Discrepancy
  Optimization}.
\newblock In \emph{UAI}, 2015.

\bibitem[Engstrom et~al.(2020)Engstrom, Ilyas, Santurkar, Tsipras, Janoos,
  Rudolph, and Madry]{engstrom2020implementation}
Logan Engstrom, Andrew Ilyas, Shibani Santurkar, Dimitris Tsipras, Firdaus
  Janoos, Larry Rudolph, and Aleksander Madry.
\newblock {Implementation Matters in Deep Policy Gradients: A Case Study on PPO
  and TRPO}.
\newblock In \emph{ICLR}, 2020.

\bibitem[Finn et~al.(2016)Finn, Christiano, Abbeel, and
  Levine]{finn2016connection}
Chelsea Finn, Paul Christiano, Pieter Abbeel, and Sergey Levine.
\newblock {A Connection Between Generative Adversarial Networks, Inverse
  Reinforcement Learning, and Energy-based Models}.
\newblock \emph{arXiv:1611.03852}, 2016.

\bibitem[Fu et~al.(2018)Fu, Luo, and Levine]{fu2018learning}
Justin Fu, Katie Luo, and Sergey Levine.
\newblock {Learning Robust Rewards with Adversarial Inverse Reinforcement
  Learning}.
\newblock In \emph{ICLR}, 2018.

\bibitem[Fu et~al.(2020)Fu, Kumar, Nachum, Tucker, and Levine]{fu2020d4rl}
Justin Fu, Aviral Kumar, Ofir Nachum, George Tucker, and Sergey Levine.
\newblock {D4RL: Datasets for Deep Data-driven Reinforcement Learning}.
\newblock \emph{arXiv:2004.07219}, 2020.

\bibitem[Fujimoto et~al.(2018)Fujimoto, Hoof, and
  Meger]{fujimoto2018addressing}
Scott Fujimoto, Herke Hoof, and David Meger.
\newblock {Addressing Function Approximation Error in Actor-critic Methods}.
\newblock In \emph{ICML}, 2018.

\bibitem[Ghasemipour et~al.(2020)Ghasemipour, Zemel, and
  Gu]{ghasemipour2020divergence}
Seyed Kamyar~Seyed Ghasemipour, Richard Zemel, and Shixiang Gu.
\newblock {A Divergence Minimization Perspective on Imitation Learning
  Methods}.
\newblock In \emph{CoRL}, 2020.

\bibitem[Goodfellow et~al.(2014)Goodfellow, Pouget-Abadie, Mirza, Xu,
  Warde-Farley, Ozair, Courville, and Bengio]{goodfellow2014generative}
Ian~J Goodfellow, Jean Pouget-Abadie, Mehdi Mirza, Bing Xu, David Warde-Farley,
  Sherjil Ozair, Aaron Courville, and Yoshua Bengio.
\newblock {Generative Adversarial Networks}.
\newblock In \emph{NeurIPS}, 2014.

\bibitem[Gretton et~al.(2012)Gretton, Borgwardt, Rasch, Sch{\"o}lkopf, and
  Smola]{gretton2012kernel}
Arthur Gretton, Karsten~M Borgwardt, Malte~J Rasch, Bernhard Sch{\"o}lkopf, and
  Alexander Smola.
\newblock {A Kernel Two-sample Test}.
\newblock \emph{JMLR}, 13\penalty0 (1):\penalty0 723--773, 2012.

\bibitem[Haarnoja et~al.(2018{\natexlab{a}})Haarnoja, Zhou, Abbeel, and
  Levine]{haarnoja2018soft}
Tuomas Haarnoja, Aurick Zhou, Pieter Abbeel, and Sergey Levine.
\newblock {Soft Actor-critic: Off-policy Maximum Entropy Deep Reinforcement
  Learning with a Stochastic Actor}.
\newblock In \emph{ICML}, 2018{\natexlab{a}}.

\bibitem[Haarnoja et~al.(2018{\natexlab{b}})Haarnoja, Zhou, Hartikainen,
  Tucker, Ha, Tan, Kumar, Zhu, Gupta, Abbeel, et~al.]{haarnoja2018softa}
Tuomas Haarnoja, Aurick Zhou, Kristian Hartikainen, George Tucker, Sehoon Ha,
  Jie Tan, Vikash Kumar, Henry Zhu, Abhishek Gupta, Pieter Abbeel, et~al.
\newblock {Soft Actor-critic Algorithms and Applications}.
\newblock \emph{arXiv:1812.05905}, 2018{\natexlab{b}}.

\bibitem[Henderson et~al.(2018)Henderson, Islam, Bachman, Pineau, Precup, and
  Meger]{henderson2018deep}
Peter Henderson, Riashat Islam, Philip Bachman, Joelle Pineau, Doina Precup,
  and David Meger.
\newblock {Deep Reinforcement Learning that Matters}.
\newblock In \emph{AAAI}, 2018.

\bibitem[Ho and Ermon(2016)]{ho2016generative}
Jonathan Ho and Stefano Ermon.
\newblock {Generative Adversarial Imitation Learning}.
\newblock In \emph{NeurIPS}, 2016.

\bibitem[Hussein et~al.(2017)Hussein, Gaber, Elyan, and
  Jayne]{hussein2017imitation}
Ahmed Hussein, Mohamed~Medhat Gaber, Eyad Elyan, and Chrisina Jayne.
\newblock {Imitation Learning: A Survey of Learning Methods}.
\newblock \emph{ACM CSUR}, 50\penalty0 (2):\penalty0 1--35, 2017.

\bibitem[Hussenot et~al.(2021)Hussenot, Andrychowicz, Vincent, Dadashi,
  Raichuk, Stafiniak, Girgin, Marinier, Momchev, Ramos,
  et~al.]{hussenot2021hyperparameter}
Leonard Hussenot, Marcin Andrychowicz, Damien Vincent, Robert Dadashi, Anton
  Raichuk, Lukasz Stafiniak, Sertan Girgin, Raphael Marinier, Nikola Momchev,
  Sabela Ramos, et~al.
\newblock {Hyperparameter Selection for Imitation Learning}.
\newblock In \emph{ICML}, 2021.

\bibitem[Jaynes(1957)]{jaynes1957information}
Edwin~T Jaynes.
\newblock {Information Theory and Statistical Mechanics}.
\newblock \emph{Phys. Rev.}, 106\penalty0 (4):\penalty0 620, 1957.

\bibitem[Jena et~al.(2020)Jena, Agrawal, and Sycara]{jena2020addressing}
Rohit Jena, Siddharth Agrawal, and Katia Sycara.
\newblock {Addressing Reward Bias in Adversarial Imitation Learning with
  Neutral Reward Functions}.
\newblock In \emph{Deep RL Workshop, NeurIPS}, 2020.

\bibitem[Kim and Park(2018)]{kim2018imitation}
Kee-Eung Kim and Hyun~Soo Park.
\newblock {Imitation Learning via Kernel Mean Embedding}.
\newblock In \emph{AAAI}, 2018.

\bibitem[Kostrikov et~al.(2019)Kostrikov, Agrawal, Dwibedi, Levine, and
  Tompson]{kostrikov2019discriminator}
Ilya Kostrikov, Kumar~Krishna Agrawal, Debidatta Dwibedi, Sergey Levine, and
  Jonathan Tompson.
\newblock {Discriminator-actor-critic: Addressing Sample Inefficiency and
  Reward Bias in Adversarial Imitation Learning}.
\newblock In \emph{ICLR}, 2019.

\bibitem[Lakshminarayanan et~al.(2017)Lakshminarayanan, Pritzel, and
  Blundell]{lakshminarayanan2017simple}
Balaji Lakshminarayanan, Alexander Pritzel, and Charles Blundell.
\newblock {Simple and Scalable Predictive Uncertainty Estimation using Deep
  Ensembles}.
\newblock In \emph{NeurIPS}, 2017.

\bibitem[Li et~al.(2015)Li, Swersky, and Zemel]{li2015generative}
Yujia Li, Kevin Swersky, and Rich Zemel.
\newblock {Generative Moment Matching Networks}.
\newblock In \emph{ICML}, 2015.

\bibitem[Lin(1992)]{lin1992self}
Long-Ji Lin.
\newblock {Self-improving Reactive Agents Based on Reinforcement Learning,
  Planning and Teaching}.
\newblock \emph{Mach. Learn.}, 8\penalty0 (3-4):\penalty0 293--321, 1992.

\bibitem[Loshchilov and Hutter(2019)]{loshchilov2019decoupled}
Ilya Loshchilov and Frank Hutter.
\newblock {Decoupled Weight Decay Regularization}.
\newblock In \emph{ICLR}, 2019.

\bibitem[Machado et~al.(2018)Machado, Bellemare, Talvitie, Veness, Hausknecht,
  and Bowling]{machado2018revisiting}
Marlos~C Machado, Marc~G Bellemare, Erik Talvitie, Joel Veness, Matthew
  Hausknecht, and Michael Bowling.
\newblock {Revisiting the Arcade Learning Environment: Evaluation Protocols and
  Open Problems for General Agents}.
\newblock \emph{JAIR}, 61:\penalty0 523--562, 2018.

\bibitem[M{\"u}ller(1997)]{muller1997integral}
Alfred M{\"u}ller.
\newblock {Integral Probability Metrics and Their Generating Classes of
  Functions}.
\newblock \emph{Adv. Appl. Probab.}, 29\penalty0 (2):\penalty0 429--443, 1997.

\bibitem[Musgrave et~al.(2020)Musgrave, Belongie, and Lim]{musgrave2020metric}
Kevin Musgrave, Serge Belongie, and Ser-Nam Lim.
\newblock {A Metric Learning Reality Check}.
\newblock In \emph{ECCV}, 2020.

\bibitem[Ng et~al.(1999)Ng, Harada, and Russell]{ng1999policy}
Andrew~Y Ng, Daishi Harada, and Stuart Russell.
\newblock {Policy Invariance Under Reward Transformations: Theory and
  Application to Reward Shaping}.
\newblock In \emph{ICML}, 1999.

\bibitem[Ng et~al.(2000)Ng, Russell, et~al.]{ng2000algorithms}
Andrew~Y Ng, Stuart~J Russell, et~al.
\newblock {Algorithms for Inverse Reinforcement Learning}.
\newblock In \emph{ICML}, 2000.

\bibitem[Oliver et~al.(2018)Oliver, Odena, Raffel, Cubuk, and
  Goodfellow]{oliver2018realistic}
Avital Oliver, Augustus Odena, Colin~A Raffel, Ekin~Dogus Cubuk, and Ian
  Goodfellow.
\newblock {Realistic Evaluation of Deep Semi-supervised Learning Algorithms}.
\newblock In \emph{NeurIPS}, 2018.

\bibitem[Orsini et~al.(2021)Orsini, Raichuk, Hussenot, Vincent, Dadashi,
  Girgin, Geist, Bachem, Pietquin, and Andrychowicz]{orsini2021matters}
Manu Orsini, Anton Raichuk, L{\'e}onard Hussenot, Damien Vincent, Robert
  Dadashi, Sertan Girgin, Matthieu Geist, Olivier Bachem, Olivier Pietquin, and
  Marcin Andrychowicz.
\newblock {What Matters for Adversarial Imitation Learning?}
\newblock In \emph{NeurIPS}, 2021.

\bibitem[Pardo et~al.(2018)Pardo, Tavakoli, Levdik, and
  Kormushev]{pardo2018time}
Fabio Pardo, Arash Tavakoli, Vitaly Levdik, and Petar Kormushev.
\newblock {Time Limits in Reinforcement Learning}.
\newblock In \emph{ICML}, 2018.

\bibitem[Paszke et~al.(2019)Paszke, Gross, Massa, Lerer, Bradbury, Chanan,
  Killeen, Lin, Gimelshein, Antiga, et~al.]{paszke2019pytorch}
Adam Paszke, Sam Gross, Francisco Massa, Adam Lerer, James Bradbury, Gregory
  Chanan, Trevor Killeen, Zeming Lin, Natalia Gimelshein, Luca Antiga, et~al.
\newblock {PyTorch: An Imperative Style, High-Performance Deep Learning
  Library}.
\newblock In \emph{NeurIPS}, 2019.

\bibitem[Piot et~al.(2014)Piot, Geist, and Pietquin]{piot2014boosted}
Bilal Piot, Matthieu Geist, and Olivier Pietquin.
\newblock {Boosted and Reward-regularized Classification for Apprenticeship
  Learning}.
\newblock In \emph{AAMAS}, 2014.

\bibitem[Pomerleau(1988)]{pomerleau1988alvinn}
Dean~A Pomerleau.
\newblock {ALVINN: An Autonomous Land Vehicle in a Neural Network}.
\newblock In \emph{NeurIPS}, 1988.

\bibitem[Raffin et~al.(2021)Raffin, Hill, Gleave, Kanervisto, Ernestus, and
  Dormann]{raffin2021stable}
Antonin Raffin, Ashley Hill, Adam Gleave, Anssi Kanervisto, Maximilian
  Ernestus, and Noah Dormann.
\newblock {Stable-baselines3: Reliable Reinforcement Learning Implementations}.
\newblock \emph{JMLR}, 22\penalty0 (1):\penalty0 12348--12355, 2021.

\bibitem[Reddy et~al.(2020)Reddy, Dragan, and Levine]{reddy2020sqil}
Siddharth Reddy, Anca~D Dragan, and Sergey Levine.
\newblock {SQIL: Imitation Learning via Reinforcement Learning with Sparse
  Rewards}.
\newblock In \emph{ICLR}, 2020.

\bibitem[Ross and Bagnell(2010)]{ross2010efficient}
St{\'e}phane Ross and Drew Bagnell.
\newblock {Efficient Reductions for Imitation Learning}.
\newblock In \emph{AISTATS}, 2010.

\bibitem[Ross et~al.(2011)Ross, Gordon, and Bagnell]{ross2011reduction}
St{\'e}phane Ross, Geoffrey Gordon, and Drew Bagnell.
\newblock {A Reduction of Imitation Learning and Structured Prediction to
  No-regret Online Learning}.
\newblock In \emph{AISTATS}, 2011.

\bibitem[Srivastava et~al.(2014)Srivastava, Hinton, Krizhevsky, Sutskever, and
  Salakhutdinov]{srivastava2014dropout}
Nitish Srivastava, Geoffrey Hinton, Alex Krizhevsky, Ilya Sutskever, and Ruslan
  Salakhutdinov.
\newblock {Dropout: A Simple Way to Prevent Neural Networks from Overfitting}.
\newblock \emph{JMLR}, 15\penalty0 (1):\penalty0 1929--1958, 2014.

\bibitem[Sutton and Barto(2018)]{sutton2018reinforcement}
Richard~S Sutton and Andrew~G Barto.
\newblock \emph{{Reinforcement Learning: An Introduction}}.
\newblock MIT Press, 2018.

\bibitem[Swamy et~al.(2021)Swamy, Choudhury, Bagnell, and Wu]{swamy2021moments}
Gokul Swamy, Sanjiban Choudhury, J~Andrew Bagnell, and Steven Wu.
\newblock {Of Moments and Matching: A Game-theoretic Framework for Closing the
  Imitation Gap}.
\newblock In \emph{ICML}, 2021.

\bibitem[Syed et~al.(2008)Syed, Bowling, and Schapire]{syed2008apprenticeship}
Umar Syed, Michael Bowling, and Robert~E Schapire.
\newblock {Apprenticeship Learning using Linear Programming}.
\newblock In \emph{ICML}, 2008.

\bibitem[Todorov et~al.(2012)Todorov, Erez, and Tassa]{todorov2012mujoco}
Emanuel Todorov, Tom Erez, and Yuval Tassa.
\newblock {MuJoCo: A Physics Engine for Model-based Control}.
\newblock In \emph{IROS}, 2012.

\bibitem[Villani(2009)]{villani2009optimal}
C{\'e}dric Villani.
\newblock \emph{{Optimal Transport: Old and New}}.
\newblock Springer, 2009.

\bibitem[Wang et~al.(2019)Wang, Ciliberto, Amadori, and
  Demiris]{wang2019random}
Ruohan Wang, Carlo Ciliberto, Pierluigi~Vito Amadori, and Yiannis Demiris.
\newblock {Random Expert Distillation: Imitation Learning via Expert Policy
  Support Estimation}.
\newblock In \emph{ICML}, 2019.

\bibitem[Williams and Peng(1991)]{williams1991function}
Ronald~J Williams and Jing Peng.
\newblock {Function Optimization using Connectionist Reinforcement Learning
  Algorithms}.
\newblock \emph{Conn. Sci.}, 3\penalty0 (3):\penalty0 241--268, 1991.

\bibitem[Xu and Denil(2021)]{xu2021positive}
Danfei Xu and Misha Denil.
\newblock {Positive-unlabeled Reward Learning}.
\newblock In \emph{CoRL}, 2021.

\bibitem[Yan et~al.(2017)Yan, Ding, Li, Wang, Xu, and Zuo]{yan2017mind}
Hongliang Yan, Yukang Ding, Peihua Li, Qilong Wang, Yong Xu, and Wangmeng Zuo.
\newblock {Mind the Class Weight Bias: Weighted Maximum Mean Discrepancy for
  Unsupervised Domain Adaptation}.
\newblock In \emph{CVPR}, 2017.

\bibitem[Ziebart et~al.(2008)Ziebart, Maas, Bagnell, and
  Dey]{ziebart2008maximum}
Brian~D Ziebart, Andrew~L Maas, J~Andrew Bagnell, and Anind~K Dey.
\newblock {Maximum Entropy Inverse Reinforcement Learning}.
\newblock In \emph{AAAI}, 2008.

\end{thebibliography}

\end{document}